\documentclass[conference]{IEEEtran}
\IEEEoverridecommandlockouts
\usepackage{cite}
\usepackage{mathtools}
\usepackage{amsmath,amssymb,amsfonts}
\usepackage{algorithmic}
\usepackage{graphicx}
\usepackage{textcomp}
\usepackage{algorithmic}
\usepackage{algorithm}
\usepackage{url}
\usepackage{color}
\def\BibTeX{{\rm B\kern-.05em{\sc i\kern-.025em b}\kern-.08em
    T\kern-.1667em\lower.7ex\hbox{E}\kern-.125emX}}
\begin{document}

\title{Identification of Probability weighted ARX models with arbitrary domains}

\author{\IEEEauthorblockN{Alessandro Brusaferri\textsuperscript{a,b}, Matteo Matteucci\textsuperscript{b}, Stefano Spinelli\textsuperscript{a,b}}
	\IEEEauthorblockA{\textit{\textsuperscript{a}CNR-Institute of Intelligent Industrial Technologies and Systems for Advanced Manufacturing, Milan, Italy} \\
		\textit{\textsuperscript{b}Politecnico di Milano - Department of Electronics, Informatics and Bioengineering, Milan, Italy}\\
		$name.surname$@stiima.cnr.it, $name.surname$@polimi.it}
	
}


\maketitle
\begin{abstract}
Hybrid system identification is a key tool to achieve reliable models of Cyber-Physical Systems from data. 
PieceWise Affine models guarantees universal approximation, local linearity and equivalence to other classes of hybrid system. Still, PWA identification is a challenging problem, requiring the concurrent solution of regression and classification tasks.
In this work, we focus on the identification of PieceWise Auto Regressive with eXogenous input models with arbitrary regions (NPWARX), thus not restricted to polyhedral domains, and characterized by discontinuous maps. 
To this end, we propose a method based on a probabilistic mixture model, where the discrete state is represented through a multinomial distribution conditioned by the input regressors. The architecture is conceived following the Mixture of Expert concept, developed within the machine learning field. To achieve nonlinear partitioning, we parametrize the discriminant function using a neural network. Then, the parameters of both the ARX submodels and the classifier are concurrently estimated by maximizing the likelihood of the overall model using Expectation Maximization. The proposed method is demonstrated on a nonlinear piece-wise problem with discontinuous maps.\\ 
\end{abstract}
\section{Introduction}

The identification of hybrid dynamics is fundamental to achieve reliable models of Cyber-Physical Systems (CPS) from data \cite{Yuan2019}. Hybrid systems provide a unified framework to represent the heterogeneous interactions between control logic/rules and the continuous dynamics of CPS, including sharp changes in operating points, plants regimes, constraints on values/variations of system inputs/outputs, etc. Several models and identification approaches have been proposed in the control community during the last decades (see e.g, \cite{GARULLI2012344},\cite{lauer_book} for detailed reviews). A major distinction is in the characterization of the discrete state evolution, which is often defined as autonomous or due to partitioning of the continuous state-input domain, leading to switching and piecewise models.

In this work, we focus on the latter, representing an important class - subject of a continuous interest - due to its universal approximation properties, local linearity and equivalence to other hybrid system modeling approaches, thus fostering analysis and deployment in advanced control applications \cite{lunze_lamnabhi-lagarrigue_2009}.

Considering input-output models, PWARX (i.e.,  PieceWise Auto Regressive systems with eXogenous input) are structured by a set of affine maps and through a partition of the continuous feature space, determining the discrete mode \cite{GARULLI2012344}.
The identification of PWARX models is a challenging problem, requiring the concurrent estimation of the sub-model dynamics and the partitioning function through intertwined  regression and classification tasks \cite{lauer_book}. Besides, as logic conditions in hybrid systems can cause sudden changes in the dynamics, non-smooth function approximations have to be addresses (e.g., using discontinuous PWARX mappings), which usually results challenging using classical nonlinear regression techniques assuming continuous patterns \cite{PAOLETTI2007242}.    

Mixed-integer programming approaches have been proposed to achieve global optimal solutions to PWARX identification (see e.g., \cite{bilevel} and references therein). As the problem is NP-hard, these approaches result affordable only for very small data sets \cite{bilevel}. 
Hence, a lot of research studies have been dedicated to approximate methods and heuristics aimed to find good sub-optional solution in reasonable time.  In this context, the most popular methods include the bounded error approach \cite{bempo_bounded}, the Bayesian approach \cite{1516255}, sparse optimization \cite{OHLSSON20131045} and the broad family of clustering-based approaches (see e.g., \cite{FERRARITRECATE2003205},\cite{NAKADA2005905},\cite{breschi}). A detailed review is reported in \cite{GARULLI2012344}.     
Despite the different techniques employed, the common workflow starts from local model estimation and clustering by relaxing the piecewise constraints, followed by partitioning through linear discrimination techniques \cite{lauer_book}. Hence, as the obtained clusters are not guaranteed to be linearly separable, a post-processing phase is usually required to address misclassifications \cite{bilevel}. 

A probabilistic generalization of PWARX (i.e., PrARX) is proposed in \cite{PrARX}, resulting in a Mixture of Expert (MoE) architecture \cite{20y}, enabling concurrent estimation of the ARXs and  parameters of the boundaries in a single optimization problem.
Then, the corresponding PWARX model is obtained using a straighforward procedure. A variational inference based technique is introduced \cite{VI_PrAXR}, supporting model structure selections in cases of small data-set size. 

Nevertheless, the vast majority of existing studies focus on piecewise models characterized by polyhedral domains in the regression space. The introduction of arbitrary regions leads to the nonlinearly piecewise models (i.e, NPWARX), a direct extension of PWARX still not deeply studied and solved in the literature \cite{lauer_NPWARX}. As a matter of fact, when the target system is characterized by these arbitrary partitions, forcing linear separability through PWARX models leads to multiple identical submodels in different regions governed by the same dynamics, while NPWARX models enable a unique representation. Besides, NPWARX supports better estimations of submodels, that would be assigned by PWARX to different modes, particularly in sub-regions with fewer samples \cite{lauer_NPWARX}. 
However, NPWARX introduce a more challenging nonlinear classification problem within the identification procedure.
To address this issue, authors in \cite{lauer_NPWARX} propose a technique based on kernel regression and Support Vector Machines.

In this work, we investigate a different approach, leveraging on tools from the probabilistic mixture field. Specifically, we extend the PrARX model by introducing a nonlinear partitioning function parametrized by a neural network. Estimation is performed through an integrated optimization, by maximizing the overall model likelihood. Then, we tackle this challenging problem through Expectation Maximization
(EM), disentangling classification and regression sub-tasks
easier to solve within each iteration.

The paper is organized as follows: Section~2 states the identification problem and details the PrARX model; Section~3 reports the developed model and EM-based estimation procedure; Section~4 describes the adopted case study and summarizes the results achieved.

\section{Background}

\subsection{PWARX model identification}
Consider discrete-time dynamical systems of the form:
\begin{equation}
  y_k = f(x_k) + e_k
\end{equation}
where $y_k, e_k\in \mathbb{R}$ are, respectively, the observed system output and error term at discrete time $k\in \mathbb{Z}$ while $f(.)$ is a deterministic mapping from the regression vector: 
\begin{equation}
x_k=[y_{k-1},...,y_{k-n_a},u^T_{k-1},...,u^T_{k-n_b}]^T
\end{equation}
$u_k \in \mathbb{R}^q$ represent exogenous inputs and $n_a, n_b \in \mathbb{N}^+$ fixed input and output lags (i.e., model orders).

PieceWise affine AutoRegressive eXogenous (PWARX) models are defined by a piecewise affine mapping as follows:
\begin{align}
\begin{dcases}
y_k &= f_1(\varphi_k)=\theta_1^T \varphi_k+ e_k, \quad  \text{if}\quad x_k \in \mathcal{X}_1\\
&...\\
y_k &= f_s(\varphi_k)=\theta_s^T \varphi_k+ e_k, \quad  \text{if}\quad x_k \in \mathcal{X}_s\\
\end{dcases}
\end{align}
where $\varphi_k \equiv [x_k \ 1]^T$ is the extended regression vector,  $S \in \mathbb{N}^+$ is the finite number of discrete state (or modes), with affine submodels parameters $\theta_s \in \mathbb{R}^{r}$  for $s \in \left\lbrace 1,...,S\right\rbrace$, with $r=n_a + q\cdot n_b+1$.
The switching mechanism between the discrete states is determined by a complete partition of the regressor domain $\mathcal{X} \subseteq \mathbb{R}^r$ in the collection of convex polyhedra (i.e, regions) $\left\lbrace \mathcal{X}_s\right\rbrace_{s=1}^S$ described by:
\begin{equation}
\mathcal{X}_s=\left\lbrace x_k \in \mathbb{R}^r: H_s \varphi_k \preceq_{[s]} 0 \right\rbrace 
\end{equation}  
where the matrices $H_s \in \mathbb{R}^{\mu_s \times r}$ represent a set of hyperplanes defining the regions by $\mu_s$ linear inequalities and '$\preceq_{[s]}$' denotes a $\mu_s$-dimensional vector whose elements can be the symbols $\leq$ and $<$.
Hence, the active discrete state $z_k$ is given by the region to which the regressor $x_k$ belongs at $k$:
\begin{equation}
z_k=s \Leftrightarrow x_k \in \mathcal{X}_s, \ s=1,...,S
\end{equation}

The general PWARX identification problem include the estimation of the ARX parameters $\left\lbrace \theta_s\right\rbrace_{s=1}^S$ and the regions $\left\lbrace \mathcal{X}_s\right\rbrace_{s=1}^S$, as well as model orders $n_a, n_b$ and the number of modes $S$, given a collection of input/output data $\mathcal{D}=\left\lbrace (x_k, y_k)\right\rbrace_{k=1}^N$.\\
In this work, we adopt the common assumption of fixed modes/orders, often estimated in practice by preliminary data analysis, cross validation and model order selection techniques \cite{lauer_book}. The investigation of automatic feature selection techniques (e.g., including dedicated regularization terms) is left to future extensions. 
Besides, we assume that in $\mathcal{D}$ all modes are sufficiently excited and inputs and outputs are bounded \cite{piga}.

\subsection{Nonlinearly PieceWise ARX model}
As introduced above, NPWARX models extend PWARX by considering  nonlinear boundaries between arbitrary regions, thus not restricted to set of hyperplanes defining polyhedral domains \cite{lauer_NPWARX}.
Formally, NPWARX are defined as in (3) while relaxing the polyhedrical assumption in (4).

\subsection{Probability weighted ARX model} 
In the PrARX model proposed in \cite{PrARX}, the deterministic partition of PWARX is replaced by probabilistic boundaries, obtained by a softmax function over a linear transformation of the extended regressor (i.e, linear gate):
\begin{align}
y_k &= f_{Pr}(x_k) + e_k, \ \ \ f_{Pr}(x_k)=\sum_{s=1}^{S}p_s \theta_s^T \varphi_k \\
p_s &=\frac{\exp \left(\eta_s^T \varphi_k \right)}{1 + \sum_{j=1}^{S-1} \exp \left(\eta_j^T \varphi_k\right)}
\end{align}
where $\eta_j\in \mathbb{R}^r$ and $p_s$ denotes the probability that $\varphi_k$ belongs to the discrete state $s$. \\
Parameters $\eta$ are directly related to the hyperplanes determining the regions.
Indeed, the PrARX model can be straightforwardly transformed to a PWARX model by:
\begin{equation}
H_s= \left[(\eta_1 - \eta_s)...(\eta_S - \eta_s) \right]^T , \ s=1,...,S
\end{equation}
Besides being a probabilistic generalization of PWARX, the PrARX represents a particular form of the MoE architecture, including linear experts and linear gates \cite{20y},\cite{Weigend1995NonlinearGE}.
Due to this simplified parametrization, the steepest descent method is adopted in \cite{PrARX} to maximize the likelihood during estimation.  

\section{Methods}

\subsection{Nonlinearly PrARX model}
We augment the PrARX model, following the Mixture of Expert approach, by replacing the linear gate with a neural network aimed to learn arbitrary region partitioning the regressor space into discrete states. Hence, we obtain a probabilistic generalization of NPWARX models, as PrARX are for PWARX \cite{VI_PrAXR}. We briefly denote it NPrARX hereinafter.

Specifically, the discrete state $z_k$, at discrete time $k\in \mathbb{Z}$, is represented by a latent categorical random variable taking values in the set $\left\lbrace 1,...,S\right\rbrace $, generated according to a $x_k$-conditioned multinomial distribution:
\begin{equation}
z_k|x_k \sim Mult(1; p(z_k=1|x_k),...,p(z_k=S|x_k))
\end{equation}
The output variable $y_k$ is generated according to a set of emission distributions, conditioned by $x_k$, selected by the active discrete mode:
\begin{equation}
y_k|z_k=s, x_k \ \sim \ p(y_k|x_k, z_k=s)
\end{equation}

Hence, data generation follows a hierarchical process, starting from the mode sampling followed by output emission, both conditioned on the regressor. 

Considering an ARX model form in each discrete state, and assuming a Gaussian distribution with zero mean and standard deviation $\sigma_s$ for $e_k$, we obtain a set of $s$ noisy linear models:
\begin{equation}
y_k|z_k=s,x_k \ \sim N(.; \theta_{s,0} + \theta_{s,1:}^T x_k, \sigma_s^2)
\end{equation}
It is worth noting that, despite the Gaussian noise assumption adopted in this work, alternative noise patterns can be considered, including specific distributions for each mode \cite{handbook_mixtures}.
\\
The overall model results in a non-Markov switching probabilistic mixture \cite{VI_PrAXR}, with semi-parametric density defined as:
\begin{align}
\begin{aligned}
p(y_k|x_k, \Theta) &= \sum_{s=1}^{S}p(z_k=s, y_k|x_k) \\
&= \sum_{s=1}^{S}p(z_k=s|x_k)p(y_k|x_k,z_k=s) \\
&= \sum_{s=1}^{S}g_s(x_k, \mathcal{W})p(y_k|x_k,\theta_{s}, \sigma_s) 
\end{aligned}
\end{align}
\\
where $\Theta=\left\lbrace \mathcal{W}, \left\lbrace \mathcal{\theta}_s, \sigma_s \right\rbrace_{s=1}^S \right\rbrace $ summarizes the parameters to be estimated and $p(y_k|x_k,\theta_{s}, \sigma_s)$ is the $s$-th mode density:
\begin{equation}
p(y_k|x_k,\theta_{s}, \sigma_s)=  \frac{1}{\sqrt{2\pi}\sigma_{s}}e^{-\frac{1}{2\sigma_{s}^2} \left(y_k - \theta_{s}^T \varphi_k \right)^2}
\end{equation}
We structure the function $g_s(x_k, \mathcal{W})$, that provides the probability of each discrete state given the regressor, as follows:
\begin{equation}
g_s(x_k, \mathcal{W})=\frac{\exp \left(\alpha_s(x_k, \mathcal{W})\right)}{1 + \sum_{j=1}^{S-1} \exp \left(\alpha_j(x_k, \mathcal{W})\right)}
\end{equation}
where the logits $\alpha_j(x_k, \mathcal{W})$ are defined using a neural network processing the regressor vector $x_k$. In the probabilistic mixture literature, this is often referred to as gate.
Besides, the parameters of one of the $S$ modes is forced to the null vector to support identifiability, as suggested in \cite{20y}. \\
While several network architectures can be adopted, depending on the specific characteristics of the application at hand, in this work we employ a feed-forward form, defined as:
\begin{equation}
\alpha_s(x_k, \mathcal{W})=\sum\limits_{l=1}^{n_h}W_{l,s}^{(2)}f_l\Bigg[\sum_{i=1}^{r-1}W_{i,l}^{(1)}x_n(i)+ W_{0,l}^{(1)}\Bigg] + W_{0,s}^{(2)} \nonumber
\end{equation}
\begin{equation}
\small f_l(x)=tanh(x)=\frac{e^x - e^{-x}}{e^x - e^{-x}}
\end{equation}
where we reported a single hidden layer of $n_h\in \mathbb{N}^+$ units to lighten notation.
It it worth noting that alternative nonlinear activation functions can be adopted and that the number of hidden units and layers constitute hyperparameters to be configured. To this end, in this work, we exploit cross-validation as detailed in section IV.

As introduced above, conditioning the gating mechanism through flexible models such as neural networks provides trainable nonlinear decision boundaries for discrete state classification in the regressor space, as opposed to the polyhedral partitioning of PrARX.  
On the other hand, such flexibility comes at the cost of a more complex optimization problem for parameters estimation \cite{handbook_mixtures}. 
To address this issues, we employ Expectation Maximization (EM), as detailed in the next section.

\subsection{Parameters estimation by Expectation Maximization}

A major feature of the Probabilistic ARX approach is the concurrent learning of classification (i.e., gating function) and regression (i.e., ARX submodels) tasks, while pursuing parameters estimation.

To this end, two major families of approaches can be followed, namely the frequentist and the Bayesian. In this work we adopt the former, leaving the investigation of the second one to future extension, e.g., through the exploitation of Bayesian Neural Networks.
Hence, we target a Maximum Likelihood Estimation (MLE) of model's parameters. 

Specifically, given a data set of i.i.d. observations $\mathcal{D}=\left\lbrace (x_k, y_k)\right\rbrace_{k=1}^N$, the likelihood function factorizes as follows: 
\begin{equation}
\mathcal{L}(\Theta)=\prod_{k=1}^{N}\left[\sum_{s=1}^{S}g_s(x_k, \mathcal{W})p(y_k|x_k,\theta_{s}, \sigma_s)\right]
\end{equation}
leading to the following objective:
\begin{equation}\label{eq:objective}
\hat{\Theta}\small{=}\underset{\Theta}{\small{argmax}}\left[\sum_{k=1}^{N} ln \left[\sum_{s=1}^{S}g_s(x_k, \mathcal{W})p(y_k|x_k,\theta_{s}, \sigma_s) \right]\right] 
\end{equation}
\\
The maximization of this function is a complex task, due to the inclusion of the hidden discrete state. This is a common issue of latent variable models, from probabilistic mixtures to Hidden Markov Models. 
In this fields, EM is typically adopted to derive the MLE. In fact, it provides a more effective alternative to the direct likelihood maximization  (e.g, though steepest descent, Newton-Raphson, etc.), specially when dealing with complex models and conditioning functions \cite{handbook_mixtures}. 

Representing a general purpose algorithm for estimation with missing data, EM has several attractive features. 
First of all, as opposed to the other methods, it has been proved that the observed log-likelihood increase at each iteration \cite{handbook_mixtures}. Moreover, it provides a natural way to tackle the challenging sum inside the logarithm form of \eqref{eq:objective}, and enables the exploitation of effective closed form solution in the maximization sub-problems during the iterations. 
%
 
Considering a set of random indicator variables assigning a generating mode $s$ to each sample $k$ in the observation set,     
$Z = \left\lbrace z_{k,s}; \ k=1,...,N, s=1,...,S \right\rbrace $,
we get the joint likelihood over the complete data (i.e, including the missing indicator variables), factorizing as follows:
\begin{equation}
\mathcal{L}_{EM}(\Theta) = \prod_{k=1}^{N}\prod_{s=1}^{S}\left[g_s(x_k, \mathcal{W})p(y_k|x_k,\theta_{s}, \sigma_s)\right]^{z_{k,s}} 
\end{equation}
thus leading to a much more tractable form:
\begin{equation}
\small ln\mathcal{L}_{EM}=\sum_{k=1}^{N} \sum_{s=1}^{S}z_{k,s}\left[ln(g_s(x_k, \mathcal{W}))+ ln(p(y_k|x_k,\theta_{s}, \sigma_s)) \right]
\end{equation}
It is worth noting that the indicator variables $z_{k,s}$ filter out all but the related discrete state in the overall model.
\\
Then, the missing indicator variables are averaged out by iteratively computing the expectation of the complete data likelihood over them $\mathbb{E}_{Z}[ln(p(X,Y,Z|\Theta))]$
\begin{align}
\begin{aligned}
\mathcal{Q}(\Theta^{i}, \Theta^{i-1})&=\sum_{k=1}^{N} \sum_{s=1}^{S}\xi_{k,s}^{ \Theta^{i-1}}ln[g_s(x_k, \mathcal{W}^{i})] \\
&+ \xi_{k,s}^{ \Theta^{i-1}} ln[p_k(y_k|x_k,\theta^{i}_{s}, \sigma_s^i)] \\
&= \mathcal{Q}^c(\mathcal{W}^i) + \sum_{s=1}^{K}  \mathcal{Q}_s^r(\theta^{i}_{s}, \sigma_s)
\end{aligned}
\end{align}
where $i=1,..,M$ represents the number of iteration of the EM algorithm. Notably, the expectation over the binary indicator variables $z_{k,s}$ assume continuous values, thus simplifying the optimization problem. $\xi_{k,s}^{ \Theta^{i-1}}$ represents the expected value of $z_{k,s}$, defined as follows:
\begin{align}
\begin{aligned}
\xi_{k,s}^{\Theta^{i-1}}&=\mathbb{E}[z_{k,s}|X,Y,\Theta^{i-1}]=p(z_k=s|x_k, y_k,\Theta^{i-1}) \\
&= \frac{p(z_k=s, y_k|x_k,\Theta^{i-1})p(y_k|x_k,\theta^{i-1}_{s}, \sigma_s^{i-1})}{p(y_k|x_k)} \\
&= \frac{g_s(x_k, \mathcal{W}^{i-1})p(y_k|x_k,\theta^{i-1}_{s},\sigma_s^{i-1})}{\sum_{j=1}^{S}g_j(x_k, \mathcal{W}^{i-1})p(y_k|x_k,\theta^{i-1}_{j},\sigma_j^{i-1})}
\end{aligned}
\end{align}
constituting the posterior obtained by the normalized product of the mode-wise likelihood times the prior $g_s(x_k,\mathcal{W})$.

The training algorithm proceed by alternating between posterior $\xi_{k,s}^{ \Theta^{i-1}}$estimation (i.e, E-step) and maximization of the $\mathcal{Q}$-function (i.e, M-step), devoted to update the estimation of $\Theta$. The former employs the parameters computed by the previous M-step, starting from an initial guess. 
    
Notably, this approach disentangles the estimation of regression (ARX) and classification (gate) parameters within the iteration, resulting in sub-problems easier to tackle. 
Specifically, the former results in a set of $S$ weighted least squared problems $\mathcal{Q}_s^r(\theta^{i}_{s}, \sigma_s)$, easily solvable by: 
\begin{equation}
\theta_{s}^{i}=\underset{\theta_{s}}{argmin} \sum_{k=1}^{N}\xi_{k,s}^{\Theta^{i-1}}   \left\lbrace -\frac{1}{2(\sigma_s^{i-1})^2} \left( y_k - \theta_{s}^T \varphi_k\right) \right\rbrace 
\end{equation}
\begin{equation} 
\theta_{s}^{i}=(\Phi^T \Xi_k \Phi)^{-1}\Phi^T \Xi_s y \ , \  \Xi_s=diag(\xi_{k,:}^{\Theta^{i-1}})
\end{equation}


The latter, i.e.,  $\mathcal{Q}^c(\mathcal{W}^i)$, constitutes a 1-of-K classification task over soft labels $\xi_{k,s}^{ \Theta^{i-1}}$, where the gate learns to approximate the posterior while the ARX sub-models compete to get responsibility over samples, gaining specialization within different regions of the regression space. During prediction, the sub-model with highest probability is selected. 
It is worth noting that, as opposed to separated clustering, such an integrated discriminant strategy foster the exploitation of network flexibility on the decision surface between the classes rather than on the overall distribution \cite{IOHMM}, often resulting in improved classification performances. 
Despite a closed form solution is not available for this problem, standard network training algorithms can be exploited in order to increase $\mathcal{Q}$, thus leading to a Generalized-EM approach which still guarantee convergence \cite{handbook_mixtures}.
Specifically, we deploy the classification task from logits, increasing numerical stability with reference to the conventional cross-entropy over softmax \cite{geron2017handson}. Then, the network weights are updated using the Adam algorithm, conceived to tackle noisy and sparse gradients \cite{adam}.

To avoid potential variance collapses \cite{Murphy}, we adopt the Maximum A Posteriori estimate proposed in \cite{Fraley}, given by:
\begin{equation} 
\sigma_s^{2^i} = \frac{v^2/S + V_s^i}{\upsilon_0 + \sum_{k=1}^{N}\xi_{k,s}^{\Theta^{i-1}}+D+2} 
\end{equation}
where $v^2=(1/N)\sum_{k=1}^{N}(y_{k}-\bar{y})^2$, $\bar{y}=(1/N)\sum_{k=1}^{N}y_{k}$, $V^i_s=\sum_{k=1}^{N}\xi_{k,s}^{\Theta^{i-1}}\left(y_k - \theta_{s}^{iT} \varphi_k \right)\left(y_k - \theta_{s}^{iT} \varphi_k \right)^T$, including the weakest prior $\upsilon_0=D+2$ with data dimension $D$. 
Since not subjected to potential collapse, an overall standard deviation parameter can be estimated by the usual MLE form:
$\sigma^{2^i} = \sum_{s=1}^{S}\sum_{k=1}^{N} \xi_{k,s}^{\Theta^{i-1}} \left(y_k - \theta_{s}^{iT} \varphi_k \right)^2/\sum_{s=1}^{S} \sum_{k=1}^{N}\xi_{k,s}^{\Theta^{i-1}}$.

The training algorithm proceed by alternating E/M-steps until convergence, tested by stopping conditions on the likelihood or maximum number of iterations.
As common for EM, several initialization are required, by sampling different values of the parameters to tackle eventual convergence to poor local minima.
As shown in the result section, we found useful experimentally to sample the initial parameters of the network from relatively large values and to initialize the sub-models biases using k-means \cite{Murphy}, to avoid modes collapse.

\section{Numerical results}


To test the proposed method, we considered a slight modification of the widely adopted case study proposed in \cite{bempo_bounded}, assumed here to introduce nonlinear partitioning requirements. Besides, it has been conceived as an extension to the simpler nonlinearly piecewise affine map estimation problem considered in \cite{lauer_NPWARX}. Specifically, the system is constituted by the following components:
\begin{align}
\begin{dcases}
\text{if} &4y_{k-1} - u_{k-1} +10 < 0 \ : \\ \nonumber
&y_k =-0.4 y_{k-1} + u_{k-1} +1.5 + e_k\\
\text{if} &4y_{k-1}-u_{k-1} +10 \geq 0 \ \text{and} \ 5y_{k-1}+u_{k-1}-6 \leq 0 \ : \\
&y_k = 0.5 y_{k-1} - u_{k-1}-0.5 + e_k\\
\text{if} &5y_{k-1}+u_{k-1}-6 > 0 \ : \\
&y_k =-0.4 y_{k-1} + u_{k-1} +1.5 + e_k
\end{dcases}
\end{align}
\\
with $e_k \sim N(0,0.2^2)$ and input sampled from a uniform distribution $u_k \sim U[-4,4]$.
The aim is to infer the same dynamics occurring in two parts of the regressor space, by reconstructing two modes. We might remark that, even if this system is defined by two affine maps, identification methods based on linear partition of the regression space would require three modes as the regions are not linearly separable. 

We generated a sequence 6000 samples. The first 5000 has been devoted to training/validation and the last 1000 represents the test set. To configure the hyperparameters, we used a cross-validation procedure, leaving 1000 samples to validation. 
%

Both the overall model architecture and the EM-based estimation algorithm have been developed in Numpy-1.18. The weighted least squares subproblems are implemented using the Linear Regression model of Scikit-learn-0.23 while the neural network using Tensorflow-2.2 API. 

By cross-validation, we found experimentally that a neural network with 10 $tanh$ hidden units is sufficient to properly fit the problem at hand. The maximum number of iterations has been set to 500, with a stopping tolerance on the log-likelihood variation of 1e-4. The Adam algorithm starts with a learning rate of 0.01, running for 3 epochs mini-batches of 100 samples in each M-iteration. The network starts from random normal weights with zero-mean. Specifically, we observed that the common weight initialization to small (i.e., approximately zero) random values frequently results in convergence to poor local solutions, where the modes collapse to a common (i.e., average) mode. To avoid this issue, we set the standard deviation of the network initializer to a high value (i.e., 10), to stimulate the system to start from initial quite different local configurations, thus limiting mode collapse. Using such configuration, we found that a maximum number of 5 random executions of the estimation algorithm are sufficient to reach good local solutions.     

To evaluate the accuracy of the ARX models reconstruction and of the estimated mode sequence, we employed \cite{PILLONETTO201621}:\\ 
\begin{equation}
\small \mathcal{F}_{\theta}= \frac{1}{S}\sum_{s=1}^{S} \left(1-\frac{\lVert \theta_s - \hat{\theta}_s \lVert}{\lVert \theta_s \lVert} \right) , \ \mathcal{F}_{s}=\frac{1}{N_{test}}\sum_{k=1}^{N_{test}} \mathbf{1}_{\hat{s}_k=s_k}
\end{equation}

namely the parameters and mode fit indexes, where $\hat{\theta}_s$, $\hat{s}_k$ indicates the estimated ARX parameters and state respectively.

As the order of the discrete states in the model do not necessarily match that of the target system, post-processing is required before evaluation \cite{handbook_mixtures}. Following \cite{PILLONETTO201621}, without loss of generality, we reordered the sub-models estimate by the Euclidean norm to the target parameters. 

The obtained estimates for the ARX submodels are reported in Table I. Statistics are computed over 100 runs of the estimation algorithm using the same hyperparameters configuration.

\begin{table}[h]
	\caption{Identified ARX parameters (100 trials)}
	\begin{center}
		\begin{tabular}{|c|c|c|c|c|c|}
			\hline
			\textbf{\textit{ }}
			& \textbf{\textit{$True$}}
			& \textbf{\textit{$Mean$}}
			& \textbf{\textit{$Std$}} \\
			\hline			 
			$\theta_1$
			& [1.5, -0.4, 1]
			& [1.503, -0.400, 0.999]
			& [2.6e-4, 7.9e-5, 8.1e-5]\\
			\hline
			$\theta_2$
			& [-0.5, 0.5, -1]
			& [-0.499, 0.499, -0.995]
			& [4.3e-4, 3.7e-4, 4.2e-5]\\
			\hline
		   	$\sigma^2$
		   & 0.2
		   & 0.201
		   & 3.1e-4\\
		   \hline	
		\end{tabular}
		\label{tab:tab3}
	\end{center}
\end{table}

Figure 1 reports the evolution of the estimated ARX parameters and covariance during the iterations of 5 consecutive runs of the training algorithm. Under the adopted configuration, convergence starting from random initial conditions have been observed in approximately 60 iterations on average. The average computation time of one iteration for this case study on an PC with CPU-i7-2.5-GHz-RAM-8Gb has been 0.42 seconds.

\begin{figure}[t]
	\includegraphics[width=1\linewidth]{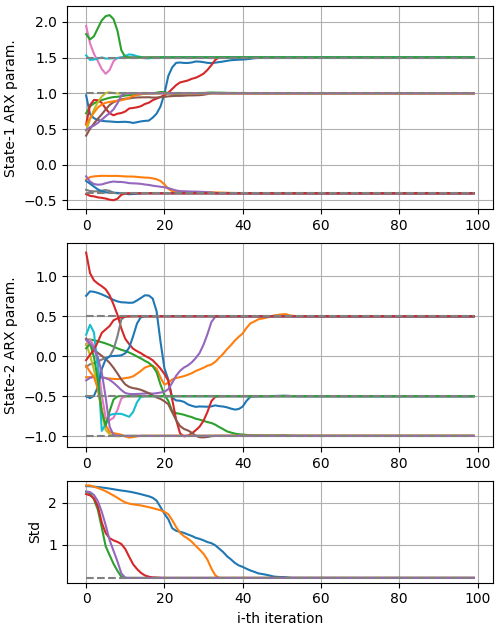}
	\caption{ARX parameters during iterations}
	\label{fig:w_hist}
\end{figure} 

The obtained performance indexes are: 
 $\mathcal{F}_{\theta}=0.997$, \  $\mathcal{F}_{s}=0.996$,
while Figure 2 reports the residuals over the test set. 

Notably, we obtained high performances over the test set. The few visible spikes in the residual plot are due to small errors in the estimated switching surface for the target discontinuous PWARX, which are in general inevitable as identification is performed over finite datasets \cite{bempo_bounded}. 
The estimated partitions of the regressor space are shown in Figure 3, where the different colors represents the states (i.e, the maximum gate network activation for each sample) and the black lines the true partitions of the target system. 

Finally, we might remark that the same estimation algorithm can be applied on a simplified discriminant function, including a linear parametrization. In this case, an EM-based estimation of the PrARX model is obtained, enabling transformation to the equivalent PWARX. However, three discrete modes would be required to enable linear separation. 
   
\begin{figure}[t]
	\includegraphics[width=1\linewidth]{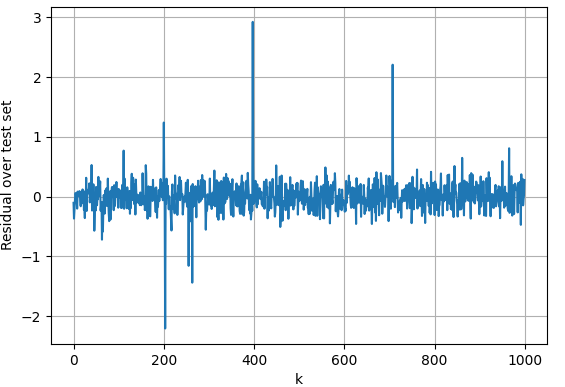}
	\caption{Residuals over the test data set}
	\label{fig:Hyb_NL}
\end{figure} 
\begin{figure}[t]
	\includegraphics[width=1\linewidth]{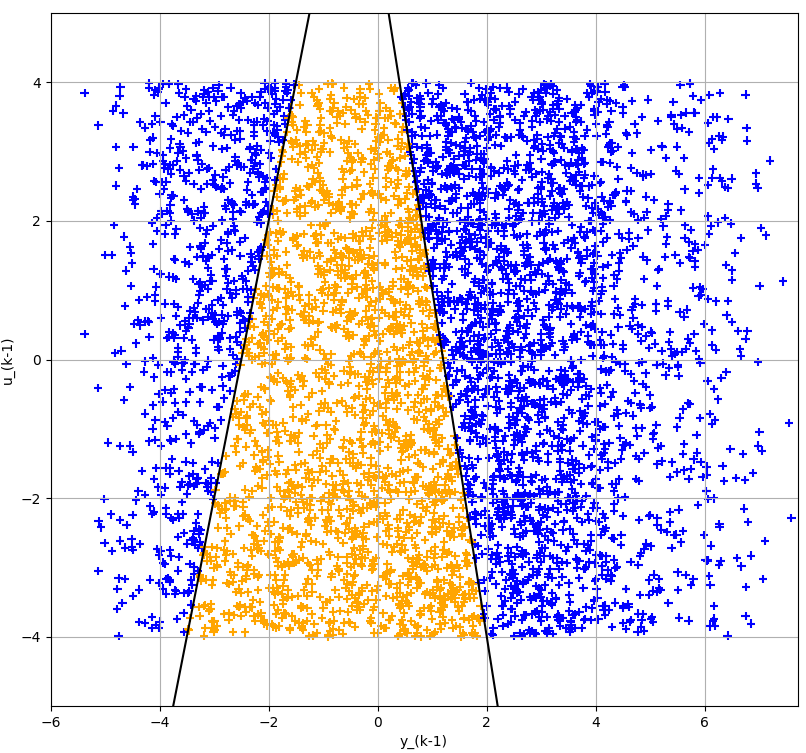}
	\caption{Discrete states classification in the regressor space}
	\label{fig:Hyb_NL}
\end{figure}

\section{Conclusion and Next Steps}
\label{Conclusion}
In this work we focused on the identification of Nonlinarly Piecewise Affine systems, characterized by arbitrary regions not restricted to polyhedral partitions.
To this end, we augmented the Probability weighted ARX  model, following a Mixture of Expert approach, by parametrizing the discriminant function through a neural network conditioned by the regressors. 
As the direct maximization of the likelihood function is particularly difficult when complex models and conditioning functions are employed, we leveraged on Expectation Maximization. Hence, ARX submodels fitting and classification tasks are disentangled within the iterations, enabling a closed form solution of the regression problems while increasing the log-likelihood at each iteration. By application to a test case we showed the capability of the proposed approach to achieve accurate parameters estimation and discrete state prediction in out of sample conditions.

Next developments will include the integration of automatic techniques for both model orders and number of modes selection, the investigation of a Bayesian inference approach and the application to a real application case. 
%

\bibliographystyle{ieeetr}
\bibliography{BLSTM}

\end{document}